\documentclass[conference]{IEEEtran}
\IEEEoverridecommandlockouts
\usepackage{cite}
\usepackage{amsmath,amssymb,amsfonts}
\usepackage{algorithmic}
\usepackage{graphicx}
\usepackage{multirow}
\usepackage{textcomp}
\usepackage{xcolor}
\def\BibTeX{{\rm B\kern-.05em{\sc i\kern-.025em b}\kern-.08em
    T\kern-.1667em\lower.7ex\hbox{E}\kern-.125emX}}
\begin{document}
\newlength\savewidth\newcommand\shline{\noalign{\global\savewidth\arrayrulewidth
  \global\arrayrulewidth 1pt}\hline\noalign{\global\arrayrulewidth\savewidth}}

\title{MV-DETR: Multi-modality indoor object detection by Multi-View DEtecton TRansformers \\
{\footnotesize \textsuperscript{*}Note: Technical Report Version}
}

\author{

\IEEEauthorblockN{1\textsuperscript{st} Zichao Dong}
\IEEEauthorblockA{\textit{UDeer.ai}\\zichao@udeer.ai}

\and

\IEEEauthorblockN{2\textsuperscript{nd} Yilin Zhang}
\IEEEauthorblockA{\textit{Zhejiang University}\\zhangyilin@zju.edu.cn}

\\

\IEEEauthorblockN{6\textsuperscript{th} Xin Zhan}
\IEEEauthorblockA{\textit{UDeer.ai} \\zhanxin@udeer.ai}

\and

\IEEEauthorblockN{3\textsuperscript{rd} Xufeng Huang}
\IEEEauthorblockA{\textit{UDeer.ai}\\xufeng@udeer.ai}

\and
\IEEEauthorblockN{4\textsuperscript{th} Hang Ji}
\IEEEauthorblockA{\textit{UDeer.ai} \\jihang@udeer.ai}
\\

\IEEEauthorblockN{7\textsuperscript{th} Junbo Chen *}
\IEEEauthorblockA{\textit{UDeer.ai} \\junbo@udeer.ai}

\and
\IEEEauthorblockN{5\textsuperscript{th} Zhan Shi}
\IEEEauthorblockA{\textit{Zhejiang University} \\zscoisini@zju.edu.cn}

\thanks{This is a draft version for now.}

}

\maketitle

\begin{abstract}  
We introduce a novel MV-DETR pipeline which is effective while efficient transformer based detection method. Given input RGBD data, we notice that there are super strong pretraining weights for RGB data while less effective works for depth related data. First and foremost , we argue that geometry and texture cues are both of vital importance while could be encoded separately. Secondly, we find that visual texture feature is relatively hard to extract compared with geometry feature in 3d space. Unfortunately, single RGBD dataset with thousands of data is not enough for training an discriminating filter for visual texture feature extraction. Last but certainly not the least, we designed a lightweight VG module consists of a visual textual encoder, a geometry encoder and a VG connector. 

Compared with previous state of the art works like V-DETR \cite{v-detr}, gains from pretrained visual encoder could be seen. Extensive experiments on ScanNetV2 dataset shows the effectiveness of our method. It is worth mentioned that our method achieve 78\% AP which create new state of the art on ScanNetv2 benchmark.

\end{abstract}


\section{Introduction}
Object detection is an important task for perception especially for autonomous driving and embodied AI. However, object detection in 3d space with multi-modality input is a hot topic in nowadays. As for the reasons, on the one hand detection result in 3d space could be used by downstream modules like navigation and planing directly. Detection results in 2d image space have to be transformed to 3d space by view transformation algorithms by contrast, which would inevitable bring accuracy dropping. On the other hand, multi-modality input would provide more information, which is definitely helpful for object detection task.  

Compared with camera or lidar, depth camera could provide sufficient and accurate information including geometry and visual context. As a result of that, depth camera is common used for indoor perception task such as object detection and segmentation. However, with RGBD data as input from raw sensor, most previous work jointly extract feature from RGBD without benefit from pretraining from modern visual foundation models. In our work, we propose a simple yet effective VG module to exact geometry and visual texture feature in parallel.  

Our contributions can be summarized as follows: 

1. Inherit modern strong visual pretraining weights to obtain discriminantive visual texture feature.  

2. A simple fusion layer to associate visual and geometry feature for each point. 

3. New state of the art result on ScanNetV2 object detection benchmark.

\section{RELATED WORK}
\subsection{Multi-modality object detection }
In the field of 3D perception, point cloud-based methods have traditionally been the dominant approach. These methods can be broadly categorized into point-based and voxel-based techniques. Point-based methods, exemplified by PointNet++ and PTV3 \cite{wu2024pointtransformerv3simpler}, effectively extract raw geometric features from point clouds but are computationally intensive due to the need for identifying neighboring relationships. This approach is particularly well-suited for indoor 3D perception tasks, where detailed geometric information is crucial.
Voxel-based methods, on the other hand, discretize the 3D space into a grid of voxels, enabling structured data processing. However, these methods are constrained by the computational demands of voxelization, often necessitating the Bird’s Eye View (BEV) assumption. This assumption simplifies 3D perception by projecting the 3D space into a 2D plane, where objects share the same x and y coordinates, making it suitable for outdoor applications.   

While point cloud-based methods excel in capturing the geometric properties of objects, they lag behind vision-based methods in semantic differentiation. Vision-based approaches, particularly those using camera data, have made significant strides in transforming 2D image features into 3D space through BEV representation. For instance, BEVFormer \cite{li2022bevformerlearningbirdseyeviewrepresentation} integrates spatial and temporal features via grid-shaped BEV queries, and BEVDepth \cite{li2022bevdepthacquisitionreliabledepth} introduces a camera-aware depth estimation module to improve 3D detection accuracy. Additionally, sparse query-based paradigms leverage global object queries derived from representative data, enabling efficient 3D bounding box prediction during inference with minimal computational overhead.
Recent advancements have explored the fusion of point cloud and camera data to leverage the strengths of both modalities. BEV-based methods are particularly promising, as they efficiently combine camera and LiDAR data into a unified BEV representation using 2D convolutions for various 3D perception tasks. However, these methods, especially those employing LSS \cite{philion2020liftsplatshootencoding} (Lift-Splat-Shoot) projection, are challenged by calibration difficulties and the ill-posed nature of the inverse problem.   
\subsection{BEVFusion}
BEVFusion \cite{liang2022bevfusionsimplerobustlidarcamera} is an efficient multi-task multi-sensor fusion framework for 3D perception in autonomous driving. It innovatively unifies camera and LiDAR features in the shared bird's-eye view (BEV) space, preserving both geometric and semantic information. The key components include an optimized BEV pooling operator that accelerates the camera-to-BEV transformation, and a convolutional BEV encoder that extracts task-agnostic features for downstream task-specific heads. BEVFusion sets a new benchmark in multi-sensor fusion for autonomous driving, providing a comprehensive solution that balances efficiency and performance by effectively integrating multi-modal data into a unified representation for enhanced 3D perception.

\subsection{View Transformation}
The generation of colored point clouds from RGB-D images is a fundamental technique in computer vision and 3D reconstruction. This process involves converting depth information and corresponding RGB values into a comprehensive 3D representation. The intrinsic parameters of the camera, such as focal length and principal point, are crucial for mapping the depth image pixels to 3D coordinates accurately. Each pixel in the depth image is transformed into a 3D point, and its corresponding color information is extracted from the RGB image. The resulting point cloud not only captures the spatial structure but also preserves the color information, enabling more detailed and realistic 3D models.

\subsection{PointNet++}
PointNet++ \cite{qi2017pointnetdeephierarchicalfeature} extends PointNet \cite{qi2017pointnetdeeplearningpoint} to capture local structures in point sets by introducing a hierarchical feature learning framework. It applies PointNet recursively on nested partitions of the input point set, learning features at increasing scales. The architecture is composed of set abstraction levels that group points into overlapping local regions and encode them using PointNet. PointNet++ demonstrates significant improvements in capturing fine-grained geometric details and local structures over its predecessor, making it a pivotal advancement in the field of 3D point cloud processing.

\subsection{EfficientViT}
Camera-based perception tasks provide relatively accurate insights into the 3D world, particularly when models are equipped with high-resolution dense prediction capabilities. However, the challenge lies in the significant computational resources and processing time required to achieve such high-resolution predictions.    

Recent advancements in vision transformers have addressed these challenges by optimizing for efficiency without sacrificing performance. One notable development is EfficientViT \cite{liu2023efficientvitmemoryefficientvision}, a new family of vision transformer models specifically designed for efficient high-resolution dense prediction. EfficientViT introduces a novel multi-scale linear attention module that supports a global receptive field and multi-scale learning while maintaining hardware-efficient operations. This innovation allows for more scalable and effective processing in high-resolution tasks, making it a promising approach in the ongoing evolution of camera-based 3D perception.

\subsection{V-DETR}
Recent advancements in 3D object detection have seen the application of transformer-based architectures to point cloud data. A notable contribution in this field is V-DETR \cite{v-detr}, which introduces a novel 3D Vertex Relative Position Encoding (3DV-RPE) scheme to enhance locality modeling in sparse 3D data. Shen et al. \cite{v-detr} demonstrate that a direct application of DETR \cite{detr} to 3D detection yields suboptimal results due to the model's inability to learn accurate inductive biases from limited training data. To address this limitation, they propose 3DV-RPE, which computes position encodings for points based on their relative positions to predicted 3D bounding box vertices. This approach provides explicit guidance for the attention mechanism to focus on relevant local regions.Notably, V-DETR achieves these gains without relying on complex operations like voxel expansion, which are common in many existing approaches \cite{cagroup3d, fcaf3d}. This suggests the potential of DETR-based methods to unify indoor and outdoor 3D detection architectures, a long-standing challenge in the field.

\section{METHOD}
\subsection{Overview}
\begin{figure*}[htbp]
    \centering
    \includegraphics[width=16cm]{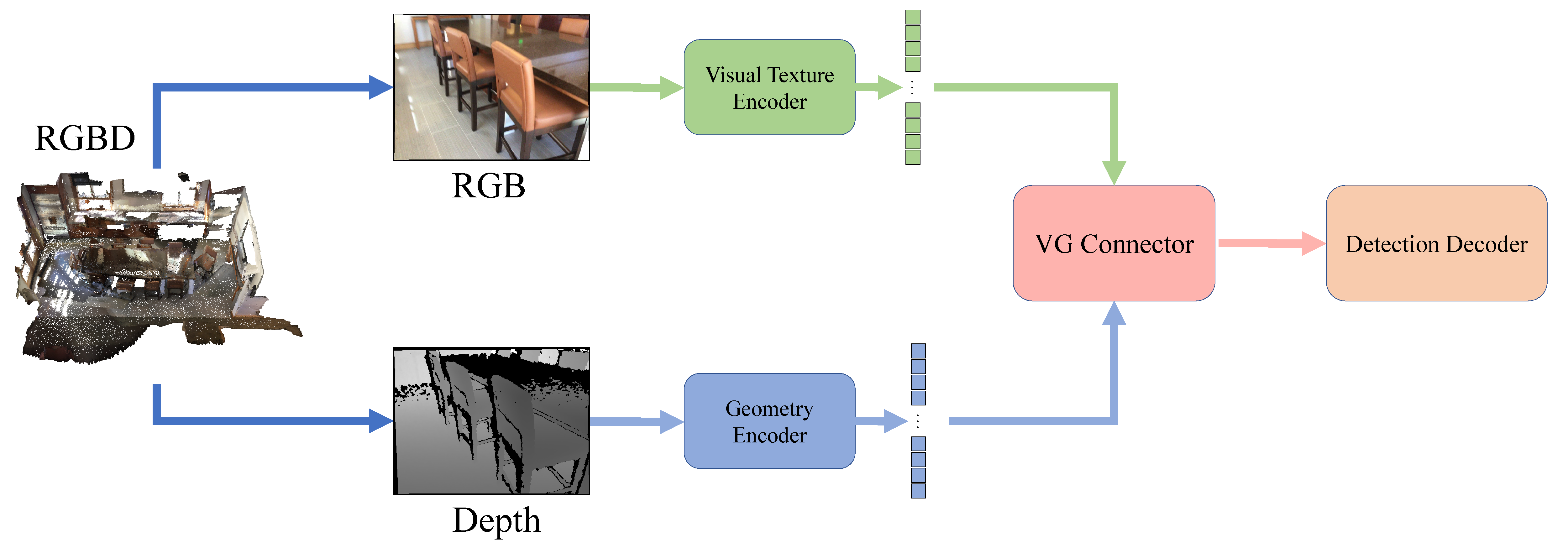}
    \caption{\small \textbf{Pipeline of MV-DETR}. The MV-DETR is mainly constructed by four main components: geometry encoder, visual texture encoder, connector and detection decoder. }
    \label{fig:model}
\end{figure*}

Our MV-DETR is mainly constructed by four main components: geometry encoder, visual texture encoder, connector and detection decoder. For short, two point embedding module will exact featur from 3d space and perspective view separately, and then be fed in detection module by a fusion layer as connector. Below section would depict them in detail. The pipeline of MV-DETR is shown in Fig.~\ref{fig:model}. Details and experiments are illustrated in below sections. 

\subsection{Geometry encoder}
Similar to PointNet \cite{qi2017pointnetdeeplearningpoint} and PointNet++ \cite{qi2017pointnetdeephierarchicalfeature}, a point level encoder in 3d space is used as our geometry encoder. It is worth mentioning that we only use xyz as input for this module. The above mentioned xyz is converted by depth and camera intrinsic in camera coordination. Firstly, we use a point embedding layer to project xyz to a single point feature with dimension d. After that, each point will communicate with his k nearest neighbours in Euclidean space by MLP. 

In our design, this layer aims to exact low level geometry features like shape while maintaining accurate 3d positional information. 

\begin{equation}
\begin{split}
\begin{pmatrix}
X \\
Y \\
Z
\end{pmatrix}= Depth(u, v) \cdot K^{-1} 
\begin{pmatrix}
u \\
v \\
1
\end{pmatrix}
\end{split}
\end{equation}

Where Depth represents the depth map, and K denotes the camera's intrinsic matrix.

\subsection{Visual texture encoder}
Apart from geometry feature, visual texture feature is undoubtedly useful for object recognition. Further, benefit from large image pretrain datasets like ImageNet \cite{russakovsky2015imagenetlargescalevisual} and ADE20k \cite{zhou2018semanticunderstandingscenesade20k}, visual texture patterns could be well learned compared with naive train from scratch. It is common sense that object could be recalled easier in RGB domain even in zero-shot manner like CLIP \cite{radford2021learningtransferablevisualmodels}. We use EfficientViT \cite{liu2023efficientvitmemoryefficientvision} block to as our implementation while borrowing pretrain from ADE20k segmentation dataset.

\begin{figure}[htbp]
   \centering
   \includegraphics[width=\columnwidth]{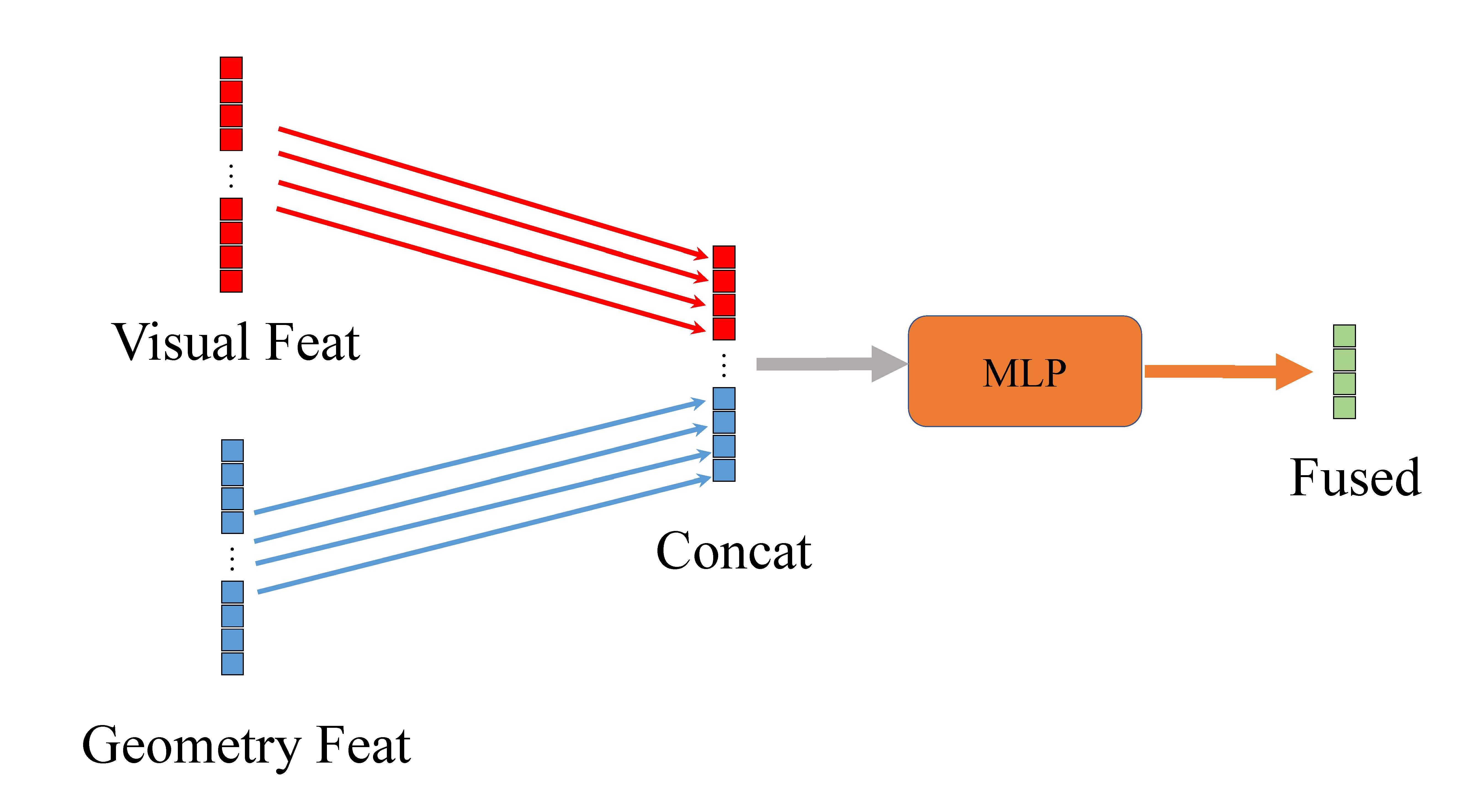}
   \caption{\small \textbf{Components of VG Connector}. We only use simple linear layer as adaptor to fuse feature from multiple domain.}
   \label{fig:fusion}
\end{figure}
\subsection{VG Connector}

For shot, VG stands for visual and geometry separately. Our connector is similar to previous multi-modality fusion based model like BEVFusion \cite{liang2022bevfusionsimplerobustlidarcamera}. Taking point-level geometry feature and visual texture feature as input, the above two features would be concatenated in channel dimension and then fed into a MLP. Activation function and normalization are also conducted. In our experiment, we found that simple MLP layer is strong enough for geometry and visual fusion task.

\subsection{Decoder for object detection}
Our method employs a decoder architecture that aligns with the approach used in V-DETR for 3D object detection. Similar to V-DETR, our framework utilizes a set of learnable 3D object queries, comprising both content and position queries, which are processed through multiple layers of a Transformer decoder to refine object representations progressively. Furthermore, we integrate the 3D Vertex Relative Position Encoding (3DV-RPE) \cite{v-detr} within the cross-attention mechanism of each decoder layer. This encoding calculates the relative position of each point concerning the vertices of the predicted 3D bounding boxes, thereby guiding the attention mechanism to focus on pertinent local regions. The 3DV-RPE is computed in a canonical object space, ensuring consistent encoding irrespective of object orientation.

Additionally, our method incorporates a lightweight feed-forward network (FFN) to predict initial 3D bounding boxes, which are subsequently refined iteratively through the decoder layers. This design enables our model to effectively capture locality in sparse 3D point clouds, addressing the challenges inherent in applying the naive DETR approach to 3D detection tasks. Ultimately, the decoder output is utilized to predict the final 3D bounding boxes and object classifications.

\section{Experiment}

\subsection{Dataset}
We validate the proposed MV-DETR on ScanNetV2 dataset.

\subsubsection{ScanNetv2}
ScanNetv2 has a total of 1613 indoor scenes, of which 1201 are used for training, 312 for validation, and 100 for testing. It consists of 3D meshes recovered from RGB-D videos captured in various indoor scenes. It has about 12K training meshes and 312 validation meshes, each annotated with semantic and instance segmentation masks for around 18 classes of objects. We follow \cite{qi2019deep} to extract the point clouds from the meshes.

\subsection{Implementation Details}
\subsubsection{Geometry encoder} 
In geometry encoder, each point is projected as a 16 dimension feature vector. Further, we set k=5 in k-nearest neighbour searching step. The KNN communication layer would be repeated twice in our MV-DETR. GELU \cite{lee2023geluactivationfunctiondeep} is used as our activation function while BN \cite{ioffe2015batchnormalizationacceleratingdeep} as normalization function. 

\subsubsection{Visual texture encoder} 
For efficiency, we use EfficientViT-b0 version as our base architecture, which is pretrained on ADE20k dataset. Notably, we only use the first two blocks to extract texture feature. Each point would be projected as a 16 dimension feature vector after this layer. 

\subsubsection{VG connector} 
As mentioned above, both visual texture module and geometry encoder module would output a 16 dimension feature vector for each point. We first concatenate them as a 32 dimension feature vector. A MLP layer with 32 perceptrons is followed. 

\subsection{Quantitive Evaluation}
The quantitive results on ScanNetV2 dataset are shown in Table \ref{tab:sota_comparison}. 

\renewcommand{\arraystretch}{1.45}
\begin{table}[!t]
\centering
\label{tab:sota_comparison}
\vspace{4mm}
\footnotesize
\tabcolsep 3pt
\setlength{\tabcolsep}{12pt}
\resizebox{1.0\linewidth}{!}
{
\begin{tabular}{@{}l|cc}
\shline
\multirow{1}{*}{Method} & \multicolumn{2}{c}{ScanNetV2}\\ 
& AP$_{25}$ & AP$_{50}$\\
\shline
TR3D~\cite{rukhovich2023tr3d} & ${72.9}$ & ${59.3}$\\
Point-GCC+TR3D~\cite{fan2023point} & ${73.1}$ & ${59.6}$\\
SPGroup3D~\cite{zhu2024spgroup3d} & ${74.3}$ & ${59.6}$\\
CAGroup3D~\cite{cagroup3d} & ${75.1}$ & ${61.3}$\\
Swin3D-L+CAGroup3D~\cite{yang2023swin3d} & ${76.4}$ & ${63.2}$\\
OneFormer3D~\cite{oneformer3d} & ${76.9}$ & ${65.3}$\\
V-DETR~\cite{v-detr} & ${77.8}$ & ${65.9}$ \\
MV-DETR(Ours)~ & ${78.0}$ & ${65.8}$ \\
\shline

\end{tabular}
}
\vspace{4mm}
\caption{\small{Comparison with the state-of-the-art on ScanNetV2.}
}
\vspace{-5mm}
\label{tab:sota_comparison}
\end{table}


\clearpage

\bibliography{references}
\bibliographystyle{plain}


\end{document}